\pdfoutput=1
\documentclass[11pt]{article}

\usepackage{acl}

\usepackage{times}
\usepackage{latexsym}

\usepackage[T1]{fontenc}
\usepackage[utf8]{inputenc}

\usepackage{microtype}

\usepackage{subfigure}
\usepackage{xspace}
\usepackage{graphicx}
\usepackage{textcomp}
\usepackage{booktabs}
\usepackage{amsmath}
\usepackage{multirow}
\usepackage{multicol}
\usepackage[ruled,vlined]{algorithm2e}

\def\figref#1{Figure~\ref{fig:#1}}
\def\figlabel#1{\label{fig:#1}\label{p:#1}}

\def\tabref#1{Table~\ref{tab:#1}}

\def\tablabel#1{\label{tab:#1}\label{p:#1}}

\title{Cross-Lingual Constituency Parsing for Middle High German:\\ A Delexicalized Approach}

\author{Ercong Nie$^{1,2}$ \qquad Helmut Schmid$^{1}$ \qquad Hinrich Sch\"utze$^{1,2}$ \\
$^{1}$Center for Information and Language Processing (CIS), LMU Munich, Germany \\
$^{2}$ Munich Center for Machine Learning (MCML), Germany \\
\texttt{nie@cis.lmu.de}}

\begin{document}
\maketitle
\begin{abstract}
Constituency parsing plays a fundamental role in advancing natural language processing (NLP) tasks.
However, training an automatic syntactic analysis system for ancient languages solely relying on annotated parse data is a formidable task due to the inherent challenges in building treebanks for such languages. It demands extensive linguistic expertise, leading to a scarcity of available resources.
To overcome this hurdle, cross-lingual transfer techniques which require minimal or even no annotated data for low-resource target languages offer a promising solution.
In this study, we focus on building a constituency parser for \textbf{M}iddle \textbf{H}igh \textbf{G}erman (\textbf{MHG}) under realistic conditions, where no annotated MHG treebank is available for training. 
In our approach, we leverage the linguistic continuity and structural similarity between MHG and \textbf{M}odern \textbf{G}erman (\textbf{MG}), along with the abundance of MG treebank resources. Specifically, by employing the \emph{delexicalization} method, we train a constituency parser on MG parse datasets and perform cross-lingual transfer to MHG parsing. 
Our delexicalized constituency parser demonstrates remarkable performance on the MHG test set, achieving an F1-score of 67.3\%. It outperforms the best zero-shot cross-lingual~\footnote{As is prevalent in the realm of multilingual NLP, the term ``zero-shot cross-lingual'' in this context pertains to a transfer learning method where we finetune the model with task-specific data in a source language and test on the target language directly~\citep{sitaram-etal-2023-everything}.} baseline by a margin of 28.6\% points.
These encouraging results underscore the practicality and potential for automatic syntactic analysis in other ancient languages that face similar challenges as MHG.

\end{abstract}

\section{Introduction}
\begin{figure*}[t]
    \centering
    \includegraphics[width=.7\linewidth]{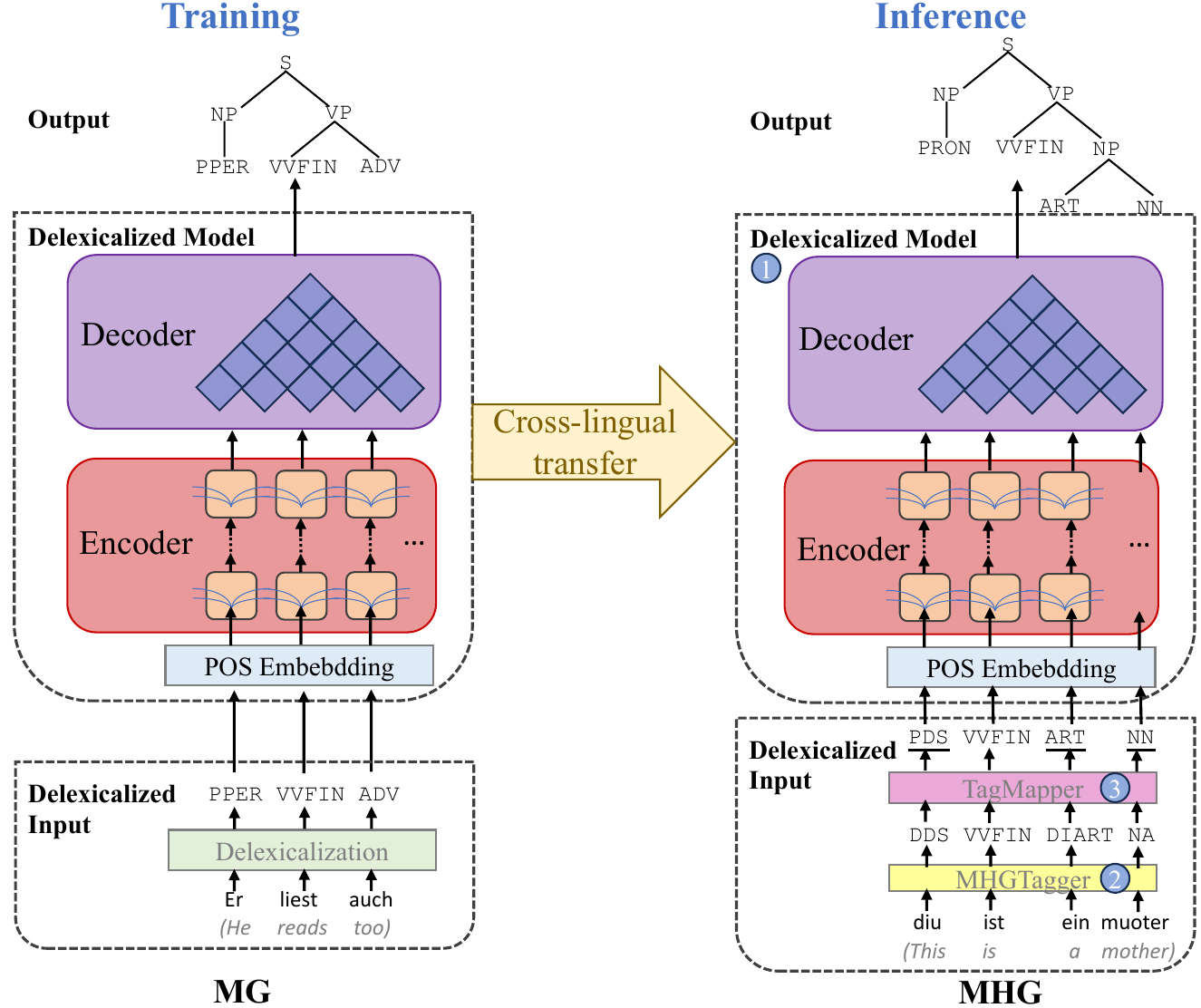}
    \caption{Overview of the cross-lingual delexicalized parsing system for MHG. In the training, the delexicalized parsing model is trained on the delexicalized MG trees. The trained parser is subsequently applied to MHG sentences. The delexicalized parsing system for MHG consists of three key modules: (1) \emph{Delexicalized parsing model} trained on delexicalized MG trees, (2) \emph{MHG POS tagger}, and (3) \emph{Tag mapper}.}
    \figlabel{method}
\end{figure*}
Constituency parsing, which involves analyzing the grammatical structure of sentences and identifying the hierarchical relationships between words, plays a crucial role in linguistic research, especially for the analysis of ancient languages that are no longer spoken.
Its significance extends beyond linguistic analysis, serving as a building block for various natural language processing (NLP) applications, such as information extraction~\citep{jiang2012information, jiang-diesner-2019-constituency}, sentiment analysis~\citep{li2020sentence}, question answering~\citep{hermjakob-2001-parsing}, etc.
However, ancient languages lack large labeled and unlabeled corpora~\citep{assael2022restoring} and treebanks suitable for parser training are seldom available. 
This scarcity of resources can be attributed to two reasons. Firstly, ancient languages usually have a dearth of digital text resources. 
Secondly, the construction of a treebank for an ancient language requires substantial linguistic expertise and manual effort.
% Nonetheless, an encouraging aspect is the continuity in the process of language evolution which gives rise to 
Nonetheless, the continuity in the process of language evolution gives rise to
linguistic similarities between ancient languages and their corresponding modern counterparts~\citep{parravicini2018continuity}. 
Cross-lingual transfer techniques~\citep{Ruder2019Neural, lauscher-etal-2020-zero} are trained on high-resource languages and require little or no annotated data from low-resource target languages. They can effectively be applied to languages with similar sentence structure and word order.
Hence, they can be a viable solution to this challenge.

In this work, we focus on building a constituency parser for Middle High German (MHG). MHG is a historical stage of the German language that was spoken between 1050 and 1350. It is the linguistic predecessor of Modern German (MG). Both languages have many similarities in word formation and grammatical features, e.g., similar word order patterns and inflectional systems~\citep{salmons2018history}.
The availability of MHG parse trees is extremely limited. The \emph{Deutsche Diachrone Baumbank (German Diachronical Treebank, DDB)}~\citep{Hirschmann2023Deutsche} comprises merely around 100 manually annotated parse trees, encompassing less than 3000 tokens. These resources are far from what is required to train an automatic syntactic analysis system, and are only suitable for use as test sets.
On the other hand, there is an abundance of treebank resources available for MG, in particular the Tiger Treebank~\citep{Smith2003ABI}.
Hence, we capitalize on the structural similarity between MHG and MG, as well as the rich MG treebank resources in order to develop a cross-lingual \emph{delexicalized} constituency parsing model that we can directly apply to MHG sentences.

In the delexicalized approach, the parsing model operates on part-of-speech (POS) sequences rather than token sequences. We accomplish this by training a cross-lingual parser using POS sequences from high-resource source languages as input. Subsequently, we utilize this trained parser to directly parse POS sequences of low-resource target languages~\citep{mcdonald-etal-2011-multi}.

In our work, we first train a delexicalized constituency parsing model on a delexicalized MG treebank. 
In order to parse MHG sentences with this model, we need to annotate them first with the POS tags used in the MG treebank.
To this end, we train a POS tagger on an MHG corpus which has been manually annotated using a POS tag set similar, but not identical to the MG tag set.
We employ a POS mapper to replace the MHG tags by the corresponding MG tags, ensuring the uniformity of the model's inputs across the two languages, which is a prerequisite of the delexicalization method.
The experimental results show that our delexicalized constituency parser substantially outperforms all other zero-shot cross-lingual parsing baselines, achieving an F1-score of 67.3\% on the MHG parse test set.

The delexicalization method is particularly well-suited for languages which (1) lack treebank resources, (2) possess sufficient annotated data for training POS taggers, and (3) exhibit syntactic similarities with a high-resource language. Our investigation of this realistic scenario shows the feasibility of automatic syntactic analysis for an ancient language.

The subsequent sections of this paper are organized as follows. In Sec.~\ref{related_work}, we discuss related work. Sec.~\ref{languages} gives an overview of our research languages and the available corpora.
The delexicalization method employed in our approach is detailed in Sec.~\ref{methods}. Sec.~\ref{experiments} describes our experimental setup, and in Sec.~\ref{results}, we analyze the results. We conclude in Sec.~\ref{conclusion}.

\section{Related Work}
\label{related_work}

\paragraph{Cross-Lingual Transfer Learning}
The fundamental principle underlying cross-lingual transfer is that the processing of source and target languages uses a shared input representation, which can be either discrete or continuous. The delexicalization method is based on a shared discrete input representation, i.e., POS tags. Other discrete representation types include glossed words~\citep{zeman-resnik-2008-cross} and grounding texts in multilingual knowledge bases~\citep{lehmann2015dbpedia}. 
Continuous cross-lingual representation spaces emerged with advancements in neural networks. Typical examples are cross-lingual word embeddings~\citep{ammar-etal-2016-many} and sentence embeddings~\citep{artetxe-schwenk-2019-massively}.

The emergence of massively multilingual transformers~\citep{devlin-etal-2019-bert, conneau-etal-2020-unsupervised}, which are jointly pretrained on multilingual corpora, introduces a novel pattern of zero-shot cross-lingual transfer learning. In this paradigm, a pretrained multilingual model is finetuned on a downstream NLP task dataset of a source language. The finetuned multilingual model is then directly applied to target language data for the same task~\citep{K2019CrossLingualAO, pmlr-v119-hu20b, liu-etal-2022-mulzdg, nie-etal-2023-crosslingualra}. 

\paragraph{Neural Constituency Parsing}
Recent advances in constituency parsing have witnessed a growing emphasis on harnessing neural network representations, making a shift from the previously prominent role of grammars, whose relevance has gradually diminished. \citet{cross-huang-2016-span} propose a span-based constituency parsing system specifically designed to leverage the powerful representation capabilities of the bidirectional long short-term memory (LSTM) networks~\citep{hochreiter1997long}. In this method, an input sentence is represented as a set of spans, and each span is assigned a score. The best-scoring parse tree is computed
using dynamic programming techniques. They combine smaller spans into larger spans until the entire sentence is covered.
Subsequently, several variations of the span-based method have been proposed, e.g.\ approaches replacing the inference algorithm with chart-based methods~\citep{stern-etal-2017-minimal}, using character-level representations instead of word-level representations~\citep{gaddy-etal-2018-whats}, and replacing LSTMs with self-attention modules~\citep{kitaev-klein-2018-constituency}. \citet{kitaev-etal-2019-multilingual} take advantage of the newly developed pretrained language models (PLMs) and use BERT~\citep{devlin-etal-2019-bert} to compute the span representations, resulting in enhanced performance. \citet{kitaev-klein-2020-tetra} improve the runtime complexity of constituency parsing to linear time by reducing parsing to tagging.

\paragraph{Cross-Lingual Constituency Parsing} 
There has been relatively limited scholarly attention dedicated to cross-lingual constituency parsing in recent studies, especially for target languages situated in low-resource settings, such as MHG. \citet{kitaev-etal-2019-multilingual} have employed the multilingual BERT model to train a single parser with parameters shared across languages. They jointly finetune the multilingual BERT on 10 languages utilizing a common BERT backbone, but the model contains distinct MLP span classifiers for each language to accommodate the different tree labels. However, their approach necessitates the availability of treebanks of all the encompassed languages as training datasets. 
\citet{kaing2021constituency} undertake a comprehensive series of experiments to validate the efficacy of delexicalization techniques for zero-shot cross-lingual constituency parsing. Additionally, their study underscores significance of typological affinity in the source language selection. We build upon these investigations and apply their findings to the zero-shot parsing of MHG within a practical contextual framework. 

\paragraph{Constituency Parsing on Historical German}
There is a notable scarcity of syntactically annotated corpora for historical German. In instances where annotated treebanks are absent, approaches such as rule-based, unsupervised, or zero-shot cross-lingual methods can be employed for constituency parsing, For instance,~\citet{chiarcos-etal-2018-analyzing} have created a rule-based shallow parser for MHG.
Recent advancements in the construction of such corpora encompass:
\begin{itemize}
    \item \emph{German Diachronical Treebank (DDB)}: a small yet syntactically deeply annotated corpus, comprising three subcorpora of different stages of German, i.e., Old High German, Middle High German and Early New High German~\citep{Hirschmann2023Deutsche}. The construction of the DDB corpus is oriented towards the Tiger Corpus~\citep{Smith2003ABI}, one of the largest German treebanks. 
    \item \emph{UP Treebank of Early New High German (ENHG)}: a syntactically annotated corpus of ENHG containing 21,432 sentences consisting of 600,569 word tokens based on the Reference Corpus of ENHG~\citep{demske2019referenzkorpus}.
    \item \emph{Corpus of Historical Low German (CHLG)}: a Penn-style treebank of Middle Low German~\citep{booth-etal-2020-penn}
\end{itemize} 
Contemporary work on historical German parsing based on previously mentioned corpora includes endeavors such as cross-dialectal parsing for ENHG based on CHLG~\citep{saap-etal-2023-parsing}.

\section{Languages and Corpora}
\label{languages}
The ancient language which we study in this paper is Middle High German (MHG). MHG and Modern German (MG) are stages of the same Germanic language family, representing different historical periods. MHG emerged during the Middle Ages in the German-speaking regions of Central Europe. It was primarily used in literary and administrative contexts and played an important role in medieval literature, including epic poems such as the \emph{Nibelungenlied} and \emph{Minnesang} (courtly love poetry)~\citep{salmons2018history}.

\paragraph{Linguistic Considerations of MHG
}MHG has a phonetic system that included a set of vowel and consonant sounds. The pronunciation and sound patterns differ from those of MG, but some MHG words are still recognizable in MG. MHG has a more complex grammatical system, such as a more extensive case system with different noun and adjective declensions. Besides, verb conjugation has more intricate forms and patterns~\citep{jones2019oxford}. In terms of orthography, the spelling and writing conventions of MHG are different from MG. For example, \emph{ü}, the umlaut of \emph{u}, is usually written \emph{iu} in MHG. 
The transition from MHG to MG was a gradual process, occurring over several centuries. MG can be considered the linguistic descendant of MHG, with linguistic changes and developments shaping the language over time.

\paragraph{MHG Corpora Resources}
During the MHG period, the amount of textual material that survives to
the present increases markedly. The \emph{Reference Corpus of Middle
  High German} (\emph{ReM})~\citep{klein2016reference} encompasses a
large collection of non-literary and non-religious texts. ReM is a
corpus of diplomatically transcribed and annotated texts of MHG with a
size of around 2 million word forms. Texts in ReM have been digitized
and richly annotated, e.g., with POS, morphological and lemma
features. The morphological annotation uses the HiTS
tag set~\citep{Dipper2013hits}, a tag set for historical German, derived
from the Stuttgart-Tübinger Tag Set (STTS) for modern German
texts~\citep{schiller1995guidelines}.  Although the ReM corpus
provides rich morphologically annotated text data for MHG, the
availability of syntactically annotated data for MHG is severely
limited, with only approximately 100 MHG parse trees included in the
DDB treebank. In contrast, the treebank resources for MG are
abundant. The Tiger Treebank~\citep{brants2002tiger}, for instance,
consists of approximately 40,000 sentences of German newspaper text,
taken from the Frankfurter Rundschau.

\section{Methods}
\label{methods}
In our work, we focus on developing a constituency parser for MHG.
In the previous section, we reviewed annotated resources available for MHG and MG. Basically, we have ample treebank resources for MG and plenty of POS-tagged texts for MHG, whereas the treebank resources for MHG are extremely limited. 
Given the resource availability for MG and MHG along with the linguistic connection between the two languages, employing a cross-lingual constituency parsing approach utilizing delexicalization proves to be an effective solution. 
As \figref{method} shows, the delexicalized model 
is trained on the delexicalized inputs of MG. In the inference stage, the delexicalized parser is directly applied to MHG POS sequences. The delexicalization method requires that MHG and MG share the same set of POS tags. 
The final constituency parser for MHG (the right side of \figref{method} comprises three modules: (1) the delexicalized parser, (2) the MHG POS tagger, and (3) the POS mapper from MHG to MG.
In the next section, we describe the delexicalized parsing system in more detail.

\subsection{Delexicalized Parser}
Our delexicalized MHG parser is based on the Berkeley neural parser (Benepar)~\citep{kitaev-klein-2018-constituency}, a span-based parser using self-attention.
As illustrated in \figref{method}, Benepar has an encoder-decoder architecture which combines a chart decoder with a sentence encoder based on self-attention.
The sentence encoder computes contextualized representations for all word positions and combines them to form span representations. From the span representations, the parser computes label scores, which are subsequently used to incrementally construct a tree
using a chart parsing algorithm~\citep{sakai-1961-syntax}.

According to \citet{kaing2021constituency}, Benepar exhibits two key features which are advantageous for cross-lingual transfer. 
Firstly, it employs a self-attentive encoder that effectively captures global context information and exhibits less sensitivity to word order.
Secondly, the parser independently scores each span without considering the label decisions of its children or parent. This means that a failure in label prediction for a certain span does not strongly impact the label prediction for other spans~\citep{gaddy-etal-2018-whats}.
Consequently, the prediction errors resulting from local syntax variations between two languages have a limited effect on the overall prediction.

While our delexicalized parser adopts the same architecture to Benepar, there exist distinctions in the inputs of the two. Specifically, Benepar is trained on parse trees with words, whereas our delexicalized parser operates on POS sequences as inputs, i.e. tree strings devoid of words. Therefore, the delexicalized version of the MG treebank is required to train the delexicalized parser. For the MHG parsing in the inference, we feed the delexicalized model with the POS sequences of MHG sentences.

\subsection{Delexicalization for MG and MHG}
\begin{figure}[t!]
\subfigure[Original MG Parse]{
\minipage{.5\textwidth}
    \centering
  \includegraphics[width=.95\linewidth]{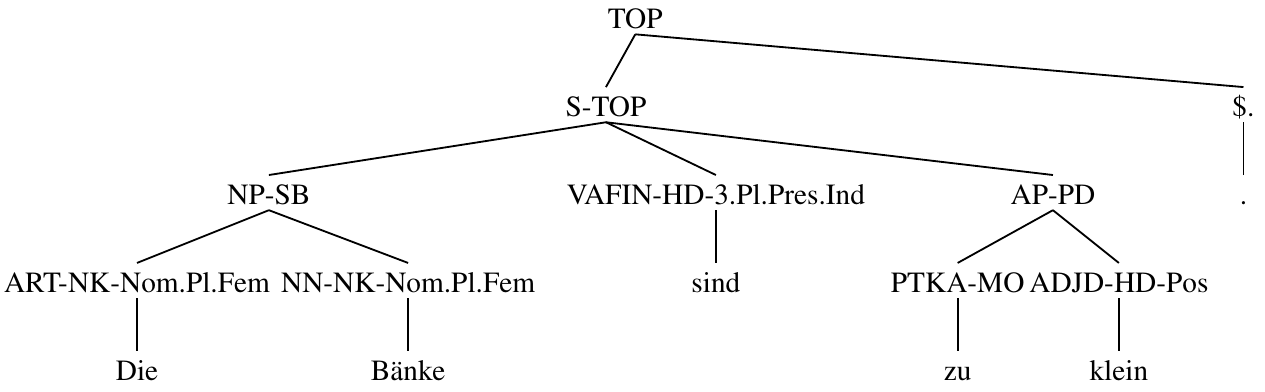}
\endminipage\hfill}
\subfigure[Delexicalized MG Parse]{
\minipage{.5\textwidth}
\centering
  \includegraphics[width=.9\linewidth]{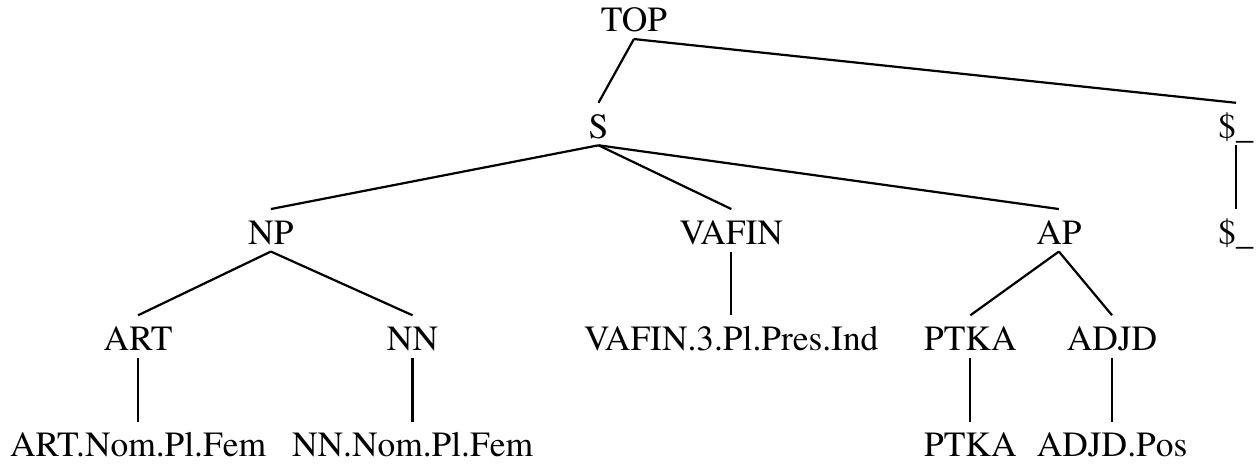}
\endminipage\hfill}
\caption{An example illustrating the delexicalization process of a MG tree.}
    \figlabel{parses}
\end{figure}

\paragraph{Delexicalization for MG}

\begin{table*}[h!]
    \centering
    \footnotesize
    \begin{tabular}{c|c|c|c|c}
    \toprule
         & \textbf{Type} & \textbf{Language} & \textbf{Size} & \textbf{Usage} \\
         \hline
       \textbf{Tiger}   & Treebank & MG & 50,474 trees & Parser training \\
       \textbf{DDB} & Treebank & MHG & 96 trees & Parser evaluation \\
       \textbf{ReM} & POS-tagged corpus & MHG & 2,269,738 tokens & POS tagger training \\
    \bottomrule
    \end{tabular}
    \caption{Overview of the datasets.}
    \tablabel{corpora}
\end{table*}

We use the Tiger Treebank to train the delexicalized parsing model on MG parse trees. The parse trees in the Tiger Treebank contain additional semantic information, such as edge labels, and special structures, such as coreference indices and trace nodes. We remove all of them during delexicalization.

In the Tiger treebank, the label of each preterminal node contains not only the POS tag, but also morphological features, such as case, number, gender.
During delexicalization, we overwrite the word at the leaf node with this extended POS tag, but only keep the POS information in the label of the preterminal node. This means that the input of our delexicalized parser contains information about morphological features.
\figref{parses} shows an example of the delexicalization for a MG sentence. As shown the edge labels, e.g., ``\texttt{NK}'' are removed and the tokens are replaced by the POS tag combined with morphological features, e.g., ``\texttt{ART.Nom.Pl.Fem}'', where ``\texttt{ART}'' (determiner) is the POS tag, and ``\texttt{Nom.Pl.Fem}'' denotes the morphological information with case being nominative, number being plural, and gender being feminine.

\paragraph{MHG POS Tagger}
For the delexicalization of MHG sentences, we need a POS tagger for MHG. We use the RNNTagger of \citet{schmid2019deep} for this purpose, which annotates MHG sentences with POS tags as well as morphological features and has been trained on the ReM corpus. RNNTagger uses deep bidirectional LSTMs with character-based word representations.

\subsection{Tag Set Mapping}

The Tiger Treebank uses the STTS tag set, whereas the MHG version of the RNNTagger and the ReM corpus on which it was trained employ the HiTS tag set. Due to this discrepancy, we cannot directly use the POS labels from RNNTagger as input to the delexicalized parser. HiTS, for example, has separate tags for definite (\texttt{DDART}) and indefinite articles (\texttt{DIART}), whereas STTS uses the tag ``\texttt{ART}'' for both of them. Since the delexicalization method demands that the source and target languages share the same tag set, we have to map the MHG tags to the MG . The small MHG treebank that we use for evaluation purposes uses STTS and requires no mapping.

\begin{table}[h]
    \centering
    \footnotesize
    \begin{tabular}{c|c}
    \toprule
       \textbf{MHG Tag}  & \textbf{MG Tag} \\
       \hline
        CARDD &	CARD  \\
        DDA	& PDAT \\
        DDART	& ART \\
        DIA	& PIAT \\
        DIART	& ART \\
        DID	& PDAT \\
        NA	& NN \\
        VAPS	& ADJD.Pos \\
        \bottomrule
    \end{tabular}
    \caption{Representative mapping pairs in the mapping dictionary.}
    \tablabel{map}
\end{table}

The mapping process involves two dimensions. Firstly, we map the morphological features of MHG to those of MG. Secondly, we map the POS tags of MHG to those of MG primarily based on a mapping dictionary. \tabref{map} shows a selected part of the POS tag mapping dictionary. 
It should be noted that our mapping is not flawless due to certain challenges. For instance, the composite word in MHG ``\emph{enerde (on earth)}'' is separated into ``\emph{auf}'' and ``\emph{Erde}'' in MG and are tagged as ``\texttt{APPR|NA}''. In the DDB treebank, such composite words are annotated with two separate tags combined with ``|'' in the DDB treebank. However, for simplification purposes, our mapping only retains the first part of the tag, leading to a loss of information.

\section{Experiments}
\label{experiments}

We begin by training Benepar on the delexicalized Tiger treebank for MG. Then we annotate the sentences of the small DDB treebank for MHG with RNNTagger and map the HiTS tags that it returns to STTS tags. Finally, we parse the POS tag sequences with the trained parser.

\subsection{Datasets}
In our experiments, we utilize the following three corpora (see also \tabref{corpora}).

\paragraph{Tiger Treebank}
The delexicalized parser is trained on the Tiger Treebank~\citep{Smith2003ABI}, which comprises a total number of 50,474 parse trees for MG. We use a version of the Tiger Treebank which has been converted to the Penn Treebank format~\citep{marcus1993building}.
We delexicalize the Tiger corpus and divide it into a training set and a development set. The first 47,474 parse trees in the Tiger corpus comprise the training set and the last 3,000 parse trees comprise the development set.

\paragraph{DDB}
The German Diachronic Treebank (DDB)~\citep{Hirschmann2023Deutsche} consists of a limited number of 100 parse trees for MHG. Due to the small data size, we utilize the DDB treebank solely for the cross-lingual evaluation of the delexicalized parser.
To prepare the DDB treebank for evaluation, we perform preprocessing steps, including converting it to the format of the Penn Treebank and removing incomplete parse trees and parse trees with mostly Latin words. We also removed numbers and periods which formed the first token of a parse tree and corrected a few more minor problems. At the end, we had 96 sentences for evaluation purposes.

\begin{table*}[!t]
    \centering
    \footnotesize
    \begin{tabular}{l|c|c|c|c|c|c|c|c}
    \toprule
         & \multicolumn{2}{c|}{\textbf{Recall}} & \multicolumn{2}{c|}{\textbf{Precision}} & \multicolumn{2}{c|}{\textbf{FScore}} & \multicolumn{2}{c}{\textbf{CM}} \\
         \cline{2-9}
         & MG & MHG & MG & MHG & MG & MHG & MG & MHG \\
         \hline
         \emph{Baselines} & & & & & & & &  \\
         \hspace{1em}Vanilla Benepar & 84.18 & 34.41 & 87.57 & 44.40 & 85.84 & 38.77 & 45.80 & 0.00 \\
         \hspace{1em}Tetra-gBERT & \textbf{86.31} & 23.20 & \textbf{88.19} & 29.53 & \textbf{87.24} & 25.98 & \textbf{51.70} & 3.12 \\
         \hspace{1em}Tetra-mBERT & 60.68 & 19.69 & 65.61 & 23.25 & 63.15 & 21.32 & 21.35 & 0.00 \\
         \hline
         \emph{Our proposed method} & & & & & & & &  \\
         \hspace{1em}Dexparser & 81.39 & \textbf{64.72} & 84.89 & \textbf{70.19} & 83.10 & \textbf{67.34} & 39.03 & \textbf{12.50} \\
         \bottomrule
    \end{tabular}
    \caption{Main results of the cross-lingual parsing transfer performance of different parsers. \textbf{CM} refers to ``complete match''. gBERT refers to the pretrained German BERT and mBERT refers to the multilingual version BERT. The best value of each column is indicated in \textbf{bold}.}
    \tablabel{results}
\end{table*}

\paragraph{ReM}
The Reference Corpus for Middle High German (ReM)~\citep{klein2016reference} is an extensive collection of texts written in MHG. This corpus encompasses approximately 2.3 million tokens and provides comprehensive linguistic annotations, including POS tags, morphological analysis, lemma features, and more. The ReM corpus has been used by \citet{schmid2019deep} to train the MHG version of his RNNTagger which annotates MHG texts with POS tags and morphological features.

\subsection{Baselines}
We evaluate the performance of our proposed delexicalized MHG parser which is based on the Benepar parser~\citep{kitaev-klein-2018-constituency}, and compare it with the cross-lingual transfer performance of the original Benepar without using the delexicalization method and
other parsing approaches that incorporate pretrained language models, which have shown promising results in various NLP tasks.

\paragraph{Vanilla Benepar}
The vanilla Benepar model is trained directly on the original training set of the Tiger Treebank for MG without delexicalization. After training, the parser is directly used to parse the MHG sentences as token sequences. This allows us to compare the performance of the delexicalized MHG parser with the vanilla Benepar model, highlighting the impact of delexicalization on cross-lingual parsing performance.

\paragraph{Tetra-Tagging with PLMs}
Tetra-tagging~\citep{kitaev-klein-2020-tetra} is a technique for reducing constituency parsing to sequence labeling. In this approach, special parsing tags are predicted in parallel using a PLM, and then merged into a parse tree. In our experiment, we use the pretrained German BERT model~\citep{chan-etal-2020-germans} and the multilingual BERT model~\citep{devlin-etal-2019-bert} available on the HuggingFace website~\citep{wolf-etal-2020-transformers}. We start by finetuning these models on the Tiger Treebank using the Tetra-tagging technique. Subsequently, we evaluate their performance on the MHG parse test set.

\subsection{Evaluation}
Following ~\citet{kitaev-klein-2018-constituency}, we use the the standard \texttt{evalb} measures \citep{sekine1997evalb, collins-1997-three} for the parser quality evaluation. \texttt{evalb} is a software tool that provides metrics to assess the accuracy and similarity of parsed sentences against reference or gold standard parse trees, including precision, recall, F1 score, and complete match.
\begin{itemize}
    \item \textbf{Precision} measures the proportion of predicted constituents in the generated parse tree which are also contained in the reference parse tree. It quantifies the accuracy of the parser in correctly identifying constituents.
    \item \textbf{Recall} measures the proportion of constituents in the reference parse tree which were predicted by the parser in the generated parse tree. It quantifies the parser's ability to generate all the constituents present in the reference parse tree.
    \item \textbf{F1 Score} is the harmonic mean of precision and recall.
    \item \textbf{Complete Match} measures the proportion of predicted parse trees which were exactly identical to the respective reference parse trees.
    \end{itemize}
As is the standard practice, the evaluation disregards POS labels and punctuation.

\subsection{Training Setup}
For training the delexicalized parser, we adopt the same hyperparameter settings as described in \citep{kitaev-klein-2018-constituency}. 
The encoder architecture consists of a character-level bidirectional LSTM neural network. We configure the encoder with a dimension of 1024, utilizing 8 layers, 8 attention heads, and a dimension of 64 for the key, query, and value. The size of the feedforward layer is set to 2048, and the character embedding dimension is 64. The batch size is set to 32, the learning rate is 5e-5, and the maximum sequence length of the encoder is 512.
We use the random seed 10 for training.
We conduct all our experiments using a server with 8 GPUs with 11GB RAM (NVIDIA GeForce GTX 1080 Ti).

\begin{table*}[!t]
    \centering
    \footnotesize
    \begin{tabular}{l|c|c|c|c}
    \toprule
         & \textbf{Recall} & \textbf{Precision} & \textbf{FScore} & \textbf{CM} \\
         \hline
        Delexicalized parser using gold tags & \textbf{66.18} & \textbf{71.17} & \textbf{68.59} & \textbf{14.58} \\
        \hspace{.5em}- \emph{using predicted tags} & 64.72 & 70.19 & 67.34 & 12.50 \\
        \hspace{1.5em}- \emph{without  mapping} & 59.16 & 68.82 & 63.63 & 7.29 \\
        \hspace{1.5em}- \emph{without morphological information} & 48.66 & 65.38 & 55.8 & 9.28 \\
    \bottomrule
    \end{tabular}
    \caption{The MHG parsing results with delexicalized parser in the ablation study.}
    \tablabel{ablation}
\end{table*}

\section{Results and Analysis}
\label{results}

\subsection{Main Results}
\tabref{results} shows the parsing performance of different cross-lingual parsers. Notably, our proposed parser attains the highest scores across all metrics for MHG, demonstrating that the delexicalized parser possesses superior cross-lingual parsing performance on MHG. Our delexicalized parser demonstrates substantial advantages in parsing MHG, achieving an impressive increase of almost 30\% points in F1 score.
Besides, it achieves comparable results on MG.
In terms of the baselines, the Vanilla Benepar and the Tetra-gBERT parser both achieve relatively high recall and precision for MG but have noticeably lower values for MHG.
The Tetra-mBERT parser exhibits lower values for both recall and precision for both MG and MHG.
It is worth noting that the parsing performance of the delexicalized model on the source language MG is surpassed by the two strong baselines, Vanilla Benepar and Tetra-gBERT. This outcome is expected as the delexicalization process diminishes the semantic information present in the input sequences. However, the trade-off of the performance loss in MG leads to a big leap in the cross-lingual parsing performance for MHG.

Our delexicalized constituency parser exhibits outstanding performance on the MHG test set, attaining an impressive F1-score of 67.3\%. This substantial improvement outperforms the best zero-shot cross-lingual baseline by a considerable margin of 28.6\%. Although there is a slight decline in the parsing performance for MG, the trade-off proves worthwhile considering the substantial gains achieved in parsing MHG. This emphasizes the effectiveness of the delexicalized approach in facilitating cross-lingual transfer and highlights its potential for parsing ancient and historical languages like MHG.

\begin{figure*}[t]
    \centering
    \includegraphics[width=.93\linewidth]{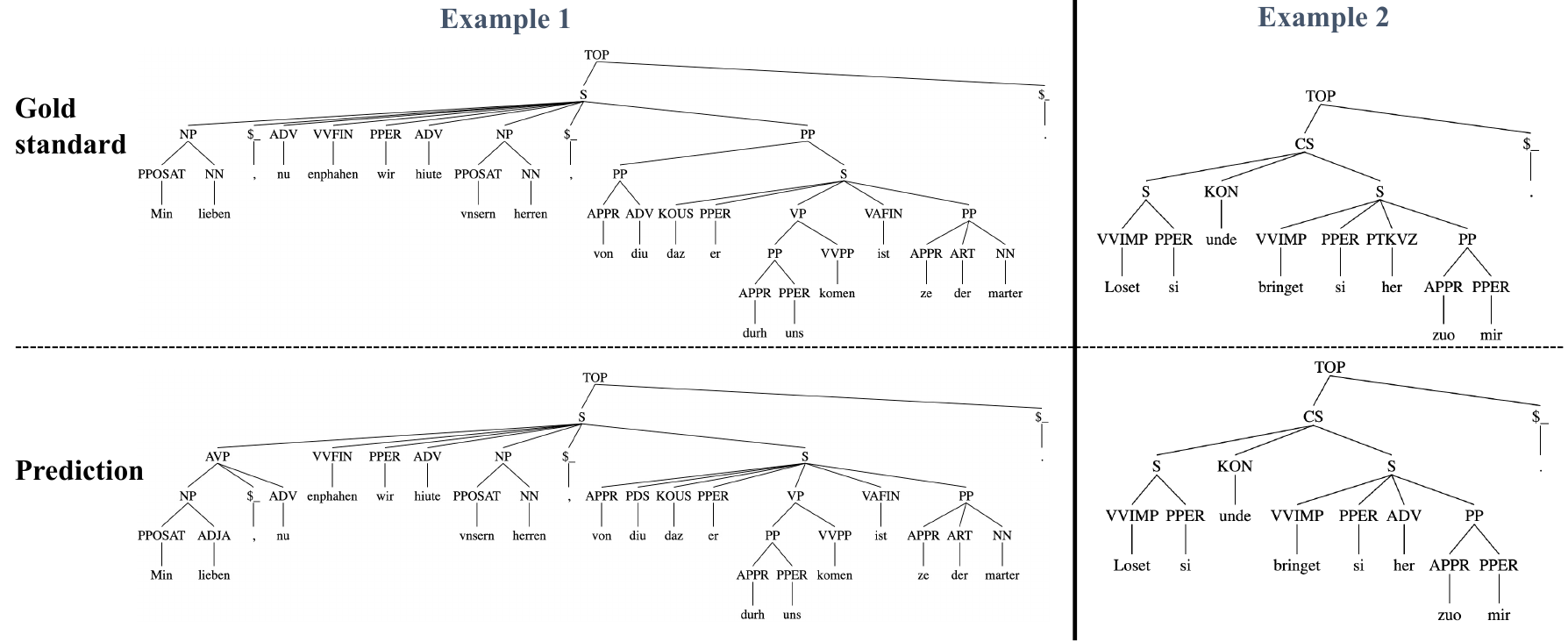}
    \caption{Two examples of the trees generated by our delexicalized parser compared to the reference parses.}
    \figlabel{cases}
\end{figure*}

\subsection{Ablation Study}
We now examine how the parsing performance changes (i) as we replace predicted POS tags with goldstandard POS tags, (ii) as we use the original HiTS tags instead of mapping them to STTS tags, and (iii) as we remove the morphological features from the parser input.
\tabref{ablation} presents the results of our ablation study.
%HS Is the default system in Table 4 really the one with goldtags and not the one with predicted tags? In other words, is "without morphological information" a system with goldstandard POS tags that lack morphological features?
% 

\paragraph{Goldstandard POS Tags}
We observe that the f-score of the delexicalized parser increases by 1.3\% points when it processes gold standard POS tag sequences instead of POS tag sequences predicted by RNNTagger. This finding underscores the quality of the POS tags predicted by RNNTagger. We loose very little performance due to POS tagging errors.
% This finding underscores the significant impact of  high-quality POS tags on improving the parsing performance. The results obtained with the predicted POS tags are remarkably close to those achieved with the gold standard tags, indicating the high quality of RNNTagger.
% This demonstrates the effectiveness and reliability of our approach in generating accurate POS tags for MHG sentences.

\paragraph{Tag Set Mapping}
\tabref{ablation} demonstrates a noticeable decline in parsing performance from 67.34\% to 43.43\% in terms of F1 score when the delexicalized MHG sequences are directly processed by the cross-lingual parser without mapping them from HiTS to STTS. This finding highlights the indispensability of  mapping from MHG to MG for maintaining satisfactory parsing performance. The results underscore the significance of aligning the tag sets between MHG and MG to ensure effective cross-lingual parsing and emphasize the necessity of this mapping process in our approach.

\paragraph{Morphological Information}
The inclusion of morphological markers provides the neural model with valuable additional information for parsing MHG sentences. In our experiments, we augment the delexicalized MHG sequences with morphological information, such as case, gender, number, and more. The outcomes of the ablation study clearly indicate that removing this morphological information from the delexicalized input sequences obviously impairs parsing performance. Specifically, this exclusion leads to a noticeable decline in the F1 score, amounting to a reduction of 11.5\%.

\subsection{Case Study}

\figref{cases} shows two MHG trees generated by our delexicalized parser and the corresponding gold standard trees for comparison.
This case study reveals that the delexicalized parser demonstrates relatively accurate predictions of constituents when compared to the reference trees, especially for short MHG sentences. Some prediction errors in constituents stem from the intricacy and the ambiguity of the MHG grammar, as exmplified by the case of ``\emph{her}'' in Example 2. From a linguistic perspective, determining whether ``\emph{her}'' functions as an adverb (\texttt{ADV}) or a separated verb prefix (\texttt{PTKVZ}) poses challenges.
However, in longer and more complex sentences, e.g., the sentence in Example 1, the parser typically maintains a high level of accuracy locally while occasionally struggling to accurately determine the overall structure of the entire sentence. Besides, the presence of noise in the ancient texts is another factor that can impact the effectiveness of the cross-lingual parsing for MHG.
Overall, the qualitative analysis provides further evidence of the effectiveness of the delexicalized parser for MHG, emphasizing its ability to accurately predict
constituents, especially in shorter sentences. While challenges may arise in handling longer and more complex sentences, the delexicalized parser showcases promising results, contributing to the advancement of MHG parsing.

\section{Conclusion}
\label{conclusion}
In conclusion, our study presents an effective cross-lingual constituency parsing approach for ancient languages, specifically focusing on the parsing of Middle High German (MHG) sentences. Through the utilization of delexicalization and and the similarities between MHG and Modern German (MG), we have developed a delexicalized parser based on the rich treebank resources of MG, which demonstrates remarkable performance in parsing MHG sentences. Our experimental results showcase the efficacy of the delexicalized approach, outperforming existing baselines and achieving substantial improvements in parsing accuracy. These findings highlight the practicality and promise of our approach for parsing historical and ancient languages, addressing the challenges posed by limited annotated data and linguistic variations.

\section*{Limitations}
One limitation of our study is the need for further improvement in the robustness of the delexicalized parsing method, particularly when applied to ancient texts. 
By addressing this limitation, we can further enhance the applicability of our approach to a wider range of ancient languages and ensure more reliable parsing results.
Besides, our proposed method is only applicable to the scenario where a POS tagger for the target language and a related language with a treebank exist.

\section*{Acknowledgements}
We extend our sincere gratitude to the anonymous reviewers for their invaluable contributions and constructive feedback that have greatly enriched the quality and scope of this paper. This work was supported by China Scholarship Council (CSC).

% Entries for the entire Anthology, followed by custom entries
\bibliography{anthology,custom}
\bibliographystyle{acl_natbib}

\appendix

% \section{Example Appendix}
% \label{sec:appendix}

\end{document}